\documentclass[conference]{IEEEtran}
\usepackage{cite}
\usepackage{amsmath,amssymb,amsfonts}
\usepackage{algorithmic}
\usepackage{graphicx}
\usepackage{textcomp}
\usepackage{csquotes}
\usepackage{caption}
\usepackage{xcolor}
\def\BibTeX{{\rm B\kern-.05em{\sc i\kern-.025em b}\kern-.08em
    T\kern-.1667em\lower.7ex\hbox{E}\kern-.125emX}}

\IEEEoverridecommandlockouts
\usepackage[breaklinks=true,bookmarks=false]{hyperref}
\hypersetup{
    colorlinks=true,
    linkcolor=cyan,
    urlcolor=blue,
    linktoc=all,
    citecolor=blue}

\usepackage{tabu}
\usepackage{multirow, makecell}

\newcommand\T{\rule{0pt}{2.9ex}}       
\newcommand\B{\rule[-1.2ex]{0pt}{0pt}} 

\begin{document}

\title{Transforming Fake News: Robust Generalisable News Classification Using Transformers}

\author{\IEEEauthorblockN{Ciara Blackledge}
\IEEEauthorblockA{\textit{School of Computing, Newcastle University} \\
Newcastle Upon Tyne, United Kingdom \\
c.blackledge1@newcastle.ac.uk}
\and
\IEEEauthorblockN{Amir Atapour-Abarghouei}
\IEEEauthorblockA{\textit{Department of Computer Science, Durham University} \\
Durham, United Kingdom \\
amir.atapour-abarghouei@durham.ac.uk}
}


\maketitle

\begin{abstract}
As online news has become increasingly popular and fake news increasingly prevalent, the ability to audit the veracity of online news content has become more important than ever. Such a task represents a binary classification challenge, for which transformers have achieved state-of-the-art results. Using the publicly available ISOT and Combined Corpus datasets, this study explores transformers' abilities to identify fake news, with particular attention given to investigating generalisation to unseen datasets with varying styles, topics and class distributions. Moreover, we explore the idea that opinion-based news articles cannot be classified as real or fake due to their subjective nature and often sensationalised language, and propose a novel two-step classification pipeline to remove such articles from both model training and the final deployed inference system. Experiments over the ISOT and Combined Corpus datasets show that transformers achieve an increase in $F_1$ scores of up to 4.9\% for out of distribution generalisation compared to baseline approaches, with a further increase of 10.1\% following the implementation of our two-step classification pipeline. To the best of our knowledge, this study is the first to investigate generalisation of transformers in this context.
\end{abstract}


\section{Introduction}
\label{sec:intro}

It has been argued that one of the great benefits of internet technology \enquote{is that it places a powerful tool of communication in the hands of the people} \cite{curran2013internet}. This democratisation of communication has allowed a shift in the way that news is written, shared and read, with the classical \enquote{top-down communication between elites and the general public} now being \enquote{subverted by horizontal communication between citizens through the internet} \cite{curran2013internet}. 

According to Ofcom’s 2020 news consumption report \cite{ofcom2020}, 65\% of adults use the internet as a news platform, while in 2016 this proportion was 41\% \cite{ofcom2015}, showing a steep increase in the availability and popularity of news online. Fletcher and Parker propose that this widespread exposure to a broad spectrum of news sources has created \enquote{a more pressing need to filter credible information} \cite{fletcher2017impact}.The issue of mistrust in online news has become prevalent in recent years, with the term \enquote{fake news} being coined in 2016 as a response to the flurry of misinformation spread surrounding the 2016 presidential election\cite{figueira2017current}. Consequently, with so many now accessing news online, verifying and auditing its veracity has become increasingly important. 

Much of the research into fake news to date falls into two categories: examining solely textual content \cite{zhou2020fake} and investigating mainly social context \cite{shu2019beyond, albahar2021hybrid}. Zhou et al. take a linguistic approach to the problem, proposing a theory driven model that explores content at \enquote{lexicon-level, syntax-level, semantic-level, and discourse-levels} \cite{zhou2020fake}. Conversely, Shu et al. pay particular attention to social context and examine the \enquote{tri-relationship, the relationship among publishers, news pieces, and users} \cite{shu2019beyond}, determining that there are correlations between these factors and the likelihood of a news story being fake, while Albahar gives equal emphasis to news content and user comments \cite{albahar2021hybrid}. Although this social context may improve the accuracy of news classification, such approaches limit detection to the point at which misinformation has already been posted, read and shared, hence only partially offering a solution to the fake news problem. This study instead focuses on identifying fake news solely from its textual content. 

Transformers have reached state-of-the-art capabilities on a range of natural language processing tasks through \enquote{relying entirely on an attention mechanism to draw global dependencies between input and output} \cite{vaswani2017attention}. They have thus \enquote{rapidly become the dominant architecture for natural language processing} \cite{wolf-etal-2020-transformers}. There is an increasing body of work exploring transformers for news classification, particularly in the context of politics (following the 2016 US presidential election) and public health (following the surge in misinformation relating to the COVID-19 pandemic) which have shown promising results when testing in distribution generalisation. However, little research has been carried out investigating the out of distribution generalisation abilities of transformers in this context.

\begin{figure*}[t!]
	\centering
	\includegraphics[width=0.9\linewidth]{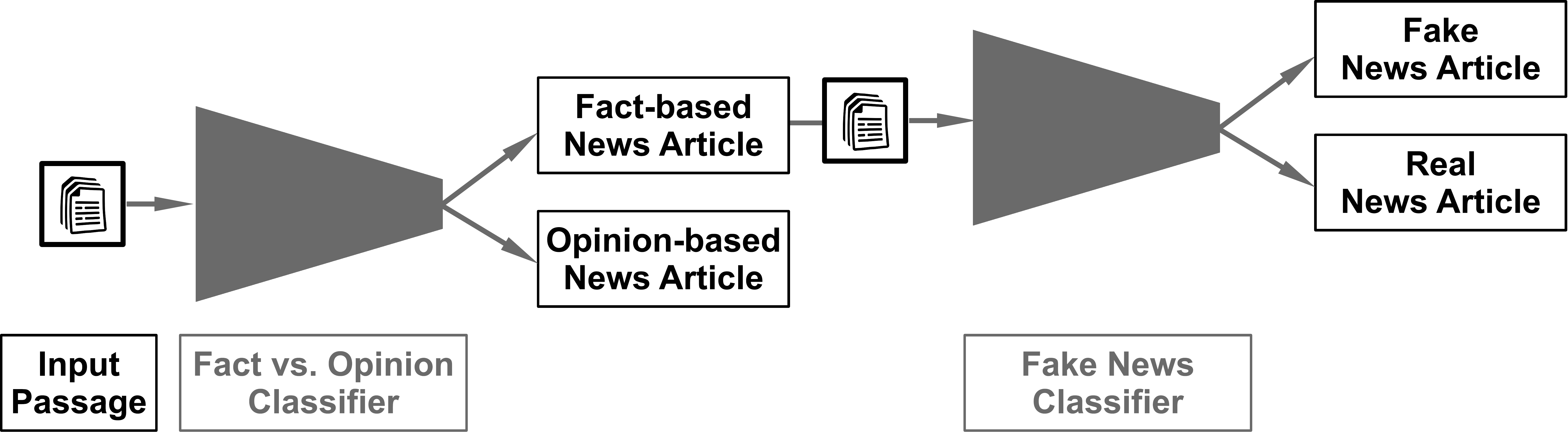}
	\captionsetup[figure]{skip=7pt}
	\captionof{figure}{Two-step opinion filtering pipeline proposed for news classification.}
	\label{fig:pipeline_diagram}
\end{figure*}

Moreover, the term \enquote{fake news} and what it constitutes is widely contested, particularly in political contexts \cite{lilleker2017evidence, graves2017anatomy}. As a result, it is likely that dataset labelling varies depending on the interpretation of fact-checkers, making these labels subjective and inconsistent. Deep learning models can be \enquote{prone to learning spurious correlations and memorizing high-frequency patterns} within data that do not generalise \cite{shorten2021text} and so these unreliable labels may harm generalisation performance. News articles that contain a large proportion of subjective information, and are therefore opinion-based, show a low consensus among fact-checkers \cite{lim2018checking} and so may make up a significant portion of potentially mislabelled samples. In order to mitigate this unreliable labelling, we propose a novel two-step classification pipeline that identifies and removes opinion-based news articles from the data used to train the final news classification model. In the proposed two-step system, the input new article is first classified as either a fact-based or opinion-based news article. Fact-based news articles are subsequently passed into a downstream classifier, which identifies whether the news article is real or fake. Fig. \ref{fig:pipeline_diagram} shows the steps involved in this classification pipeline. Further details of the proposed approach is provided in Section \ref{sec:approach}. 
To enable reproducibility, an implementation of the proposed approach is publicly available\footnote{\href{https://github.com/CiaraBee/fake_news_classification}{https://github.com/CiaraBee/fake\_news\_classification}}.

In order to assess the contributions of this work, we provide an overview of previous related work in both the fields of text classification and fake news detection in Section \ref{sec:related_work}.

\section{Related Work}
\label{sec:related_work}

We consider prior work in the context of text classification (Section \ref{sec:text_classification}) and fake news detection (Section \ref{sec:fake_news_detection}).

\subsection{Text Classification}
\label{sec:text_classification}

Text classification is one of the core tasks encompassed by natural language processing, aiming to assign a document to predefined classes \cite{dalal2011automatic}. While studies have shown machine learning techniques to achieve high performances in this field \cite{hakak2021ensemble, ott2011finding, ahmed2017detection, atapour2020rank}, the performance of such methods depends heavily on the data representation they are given. 

However, deep learning models are able to represent the world as a \enquote{nested hierarchy of concepts} through hidden layers of learning  \cite{goodfellow2016deep}. Neural networks can therefore \enquote{transform low-level features of the data into high-level abstract features} and so are generally \enquote{stronger than shallow machine learning models in feature representation} \cite{du2016overview}. Deng and Liu suggest \cite{deng2018deep} that while machine learning approaches to natural language processing tasks have reached relatively high levels of performance, they still fall short of human abilities, largely due to this \enquote{bottleneck} of feature engineering \cite{deng2018deep}. 

Minaee et al. provide a comprehensive review of transformers, machine learning and deep learning methods for text classification and find that deep learning leads to \enquote{significant improvements} across all tasks carried out \cite{minaee2021deep}. Gonzalez-Carvajal and Garrido-Merchan also show BERT based transformers to outperform traditional machine learning techniques on a range of text classification tasks, noting in particular the importance of transfer learning, attributing their strong results to pre-training \cite{gonzalez2020comparing}. 

In the context of fake news classification, which is a complex challenge even for experienced fact-checkers, models capable of understanding the \enquote{subtleties involved in conveying messages through text} \cite{thota2018fake} are necessary to tackle this nuanced problem, with transformers achieving the best results to date on such tasks.

\subsection{Fake News Detection}
\label{sec:fake_news_detection}

The consensus among researchers in the field is that there is little consensus regarding the exact definition of fake news, with Lilleker stating that this phrase has become \enquote{a catch-all term with multiple definitions} \cite{lilleker2017evidence}. As a result of this ambiguously defined problem, there has been a historic lack of labelled benchmark datasets which has \enquote{dramatically limited} statistical approaches to the fake news problem \cite{wang2017liar}. Consequently, Wang created the popular LIAR benchmark dataset, containing 12k rows of short texts labelled as one of five classes ranging from \enquote{True} to \enquote{Pants on fire}. Following this, a selection of datasets large enough for training deep learning models have been created in recent years, with Ahmed et al. releasing the ISOT fake news dataset in 2018 \cite{ahmed2017detection}, and most recently, Khan et al. releasing the Combined Corpus dataset \cite{khan_khondaker_afroz_uddin_iqbal_2021} in 2021.

\begin{table*}[!t]
	\centering
		{\tabulinesep=0mm
			\begin{tabu}{@{\extracolsep{5pt}}l c c c c@{}}
				\hline\hline
				\multirow{2}{*}{Dataset} & 
				\multirow{2}{*}{Total Rows of Data} &
				\multirow{2}{*}{Fake News} &
				\multirow{2}{*}{Real News} \\
				\T\\
\hline\hline

 ISOT Fake News Dataset & 44,898 & 21,417 & 23,481\T\B  \\ 
 \hline
 Combined Corpus Dataset & 79,548 & 40,689 & 38,859\T\B \\
 \hline
 \hline

    \end{tabu}
}
\captionsetup[table]{skip=7pt}
\captionof{table}{Details of the ISOT and Combined Corpus datasets.}
\label{table:datasets}
\end{table*}

Ahmed et al. provide initial benchmark results on the ISOT dataset showing an accuracy score of 92\% using unigram features and a Linear SVM classifier \cite{ahmed2017detection}. In another work, Ahmad et al. propose an ensemble machine learning approach of combining Decision Tree, Random Forest and Extra Tree Classifiers using bagging to aggregate their outputs and increase this accuracy to 99.8\% \cite{ahmad2020fake}.

Alameri directly compares the performance of machine learning and deep learning text classifiers on this ISOT dataset, with the Long Short Term Memory (LSTM) model outperforming others, achieving the highest performance across accuracy, precision, recall and $F_1$ metrics \cite{alameri2021comparison}. Nasir et al. propose a hybrid Convolutional Neural Network (CNN) and Recurrent Neural Network model (RNN) that \enquote{makes use of the ability of the convolutional neural networks to extract local features and of the LSTM to learn long-term dependencies}, and achieve 99\% accuracy on the ISOT dataset \cite{NASIR2021100007}, reinforcing the LSTM as a strong candidate for this task.

Aggarwal et al. investigate the performance of pre-trained BERT models on fake news detection and find that the \enquote{BERT model considerably outperforms other approaches even with minimal to no engineering of features}, concluding that transfer learning \enquote{can yield good results in the case of detection of fake news} \cite{aggarwal2020}. Radford et al. demonstrate that pre-training contributes to the strong performance of transformers on a wide variety of natural language processing fields, showing in ablation studies that its removal \enquote{hurts performance across all the tasks} \cite{radford2018improving}, and reinforcing BERT as the state-of-the-art in this context due to its \enquote{deep understanding of the language}, which is considered \enquote{necessary to detect the subtle stylistics differences in the writing of the fake articles} \cite{antounetal}. 

Khan et al. \cite{khan_khondaker_afroz_uddin_iqbal_2021} offer a benchmark study of fake news detection and find that \enquote{pre-trained BERT based models outperform the other models not only on the overall datasets but also on smaller samples}. This capacity for strong performance even on small datasets provides evidence that transformers are less prone to overfitting in this context, indicating that they may generalise well.

With vast amounts of news content constantly being published online, the ability of any fake news detection model to adapt to unseen data is critical in tackling this problem. Nasir et al. investigate generalisation to unseen news articles from a deep learning approach using their hybrid CNN-RNN model, training on the ISOT fake news dataset and testing on another, which results in \enquote{poor generalisation}, achieving results that \enquote{indicate overfitting} \cite{NASIR2021100007}. However, to the best of our knowledge there has been no research on the out of distribution generalisation abilities of transformers in this context. This study, therefore, aims to fill this gap by investigating the generalisation capabilities of transformers and proposing a two-step classification pipeline to enhance their overall performance.

\section{Data}
For the purposes of this study, which aims to investigate the applications of deep learning on the fake news problem using solely linguistic features, a large text focused dataset is needed. Marcus notes that deep learning is \enquote{data hungry} and \enquote{works best when there are thousands, millions or even billions of training examples} \cite{marcus2018deep}.  The following two datasets have therefore been identified as appropriate for this project due to their size and content. Table \ref{table:datasets} details the contents of these datasets, showing the total number of articles and proportions of real and fake news present. This table shows that the data is largely balanced, with the ISOT dataset (ISOT) containing slightly more fake news, and the Combined Corpus dataset (CC) containing slightly more real news. Further details of these two datasets are outlined in the following. 

\subsection{ISOT Fake News Dataset}
In 2017, Ahmed et al. introduced the ISOT fake news dataset \cite{ahmed2017detection} containing 45k full length articles from real world sources, with real articles collected from Reuters.com and fake articles collected from various unreliable sources. Fake news articles in this dataset were identified and sourced from Politifact.com, a not-for-profit national news organization that uses human fact-checkers to identify fake news, largely focusing on political news. This dataset therefore mainly contains news articles relating to the 2016 US presidential election and has been widely used in fake news detection studies in this field \cite{ahmed2017detection, NASIR2021100007, alameri2021comparison, hakak2021ensemble}.

\subsection{Combined Corpus Dataset}
In 2021, Khan et al. introduced the Combined Corpus dataset (CC) \cite{khan_khondaker_afroz_uddin_iqbal_2021} containing nearly 80k rows of data, with 51\% being real news and 49\% being fake news. In contrast to ISOT, the creators of this dataset actively sought out news from a wide variety of sources covering a range of topics including \enquote{national and international politics, economy, investigation, health-care, sports, entertainment, and others} \cite{khan_khondaker_afroz_uddin_iqbal_2021}, making this dataset larger in both size and scope. This dataset spans articles from 2015 to 2017 and covers a wide range of fake news types such as \enquote{hoax, satire, and propaganda} \cite{khan_khondaker_afroz_uddin_iqbal_2021}.  To the best of our knowledge, this is at present the largest publicly available fake news dataset.

\subsection{Data Preprocessing}
While this work primarily focuses on utilising solely textual features for news classification and so no feature engineering is carried out, a number of preprocessing steps have been taken to clean and prepare the ISOT and CC datasets for use. In order to preserve as many characteristic textual features as possible, all spelling and grammatical errors present within each dataset have been maintained. The focus of data preprocessing is therefore that of removing extraneous elements that may detract from these textual features.  

In this vein, all URLs, punctuation, IP addresses and links within the body of the text have been removed from the datasets via regular expressions \cite{van1995python}. Stopwords are subsequently removed from the cleaned text \cite{bird2009natural}, which is then vectorised for the LSTM model \cite{chollet2015keras} and the conventional ML models \cite{scikit-learn} used as baselines in this work. The HuggingFace library \cite{wolf-etal-2020-transformers} provides model-specific tokenizers for each transformer implementation. After completing these preprocessing steps, the data is ready for input into each model considered.

\section{Proposed Approach}
\label{sec:approach}

In order to assess the efficacy of transformers on news classification, we propose an approach comparing their performance to baseline machine and deep learning approaches. To further explain this proposed approach, details of all models implemented are outlined below, with particular attention paid to the BERT based transformers and the differences between the three architectures considered. Moreover, further detail is given regarding the two-step classification pipeline introduced in this study to further expand on the rationale and implementation for doing so. 

\subsection{Baseline Models}
Both machine and deep learning approaches are considered for baseline models with Logistic regression \cite{cox1958regression}, Naïve Bayes \cite{handnaivebayes} and Random Forest classifiers\cite{ho1995random} forming the machine learning baseline comparison.

Alameri identifies LSTMs \cite{HochSchm97} to be the best performing deep learning model for the task of news classification \cite{alameri2021comparison} and so an LSTM will therefore be used as the deep learning benchmark for this work.

BERT (Bidirectional encoder representation from transformers) \cite{devlin2018bert} builds upon the original transformer architecture \cite{vaswani2017attention} and has reached new state-of-the-art capabilities on a variety of text classification tasks \cite{devlin2018bert, sun2019utilizing, lee2020patent} by learning from a combination of pre-training and fine tuning. BERT is therefore considered \enquote{a must-have baseline} for natural language processing tasks \cite{rogers2020primer} and so is examined for its use on fake news detection in this study along with two other BERT based model types: DistilBERT \cite{sanh2019distilbert} and deBERTa \cite{he2020deberta}. 

\subsection{BERT}
BERT models \cite{devlin2018bert} differ from the original transformer \cite{vaswani2017attention} by allowing bidirectional language understanding. BERT models undergo unsupervised pretraining in a two-step process made up of masked language modelling and next sentence prediction. During masked language modelling, input tokens are randomly masked and subsequently predicted in order to obtain a \enquote{deep bidirectional representation} \cite{devlin2018bert}. This allows BERT to counter the \enquote{unidirectional constraint} \cite{gillioz2020overview} of other language models such as GPT \cite{radford2018improving} by not allowing the model to \enquote{see itself} and thus \enquote{trivially predict the next token} when learning both right to left and left to right \cite{gillioz2020overview}. The next stage of pretraining takes the form of binarised next sentence prediction where sentence A precedes sentence B 50\% of the time, allowing the model to learn the \enquote{relationship between two sentences} \cite{gillioz2020overview}. BERT models are then fine tuned by adding a classification layer and updating all parameters based on a downstream task, in this case, fake news classification. 

\subsection{DistilBERT}
DistilBERT is a distilled BERT architecture 40\% smaller than its predecessor and capable of achieving similar results while being 60\% faster during inference \cite{sanh2019distilbert}. This lightweight model is obtained through distillation, in which knowledge is transferred from a large model to a smaller, more compact counterpart \cite{hinton2015distilling}. DistilBERT has been shown to retain 97\% of the natural language understanding capabilities of its larger equivalent \cite{sanh2019distilbert}. 

\subsection{DeBERTa}
Building upon BERT, DeBERTa (Decoding-enhanced BERT with disentangled attention) proposes the addition of \enquote{a disentangled attention mechanism and an enhanced mask decoder} \cite{he2020deberta}. Dissimilarly to BERT, DeBERTa word encodings are made up of two vectors that encode both content and relative position using disentangled matrices. To enhance the masked language modelling phase of pre-training, DeBERTa \enquote{incorporates absolute word position embeddings} right before the model decodes the masked words \cite{he2020deberta}.

\subsection{Fact vs. Opinion Classification}

As discussed in Section \ref{sec:intro}, an important contribution of this work is the introduction of an additional step in the overall fake news classification pipeline to identify and remove opinion-based passages, providing a cleaner and more accurate process for detecting fake news articles.

Lim investigates fact-checking for political news and finds that the \enquote{the rate of agreement on its factual accuracy is quite low for statements in the relatively ambiguous scoring range} \cite{lim2018checking}, indicating that there is a large amount of uncertainty in classifying claims that do not directly confirm or contradict a fact, i.e., subjective or unclear claims. Furthermore, Graves identifies that many fact-checking organisations adopt the rule that there is \enquote{no way to check a statement of opinion} \cite{graves2017anatomy}, confirming that opinions cannot be fact-checked and thus exist outside the scope of news classification. 

In accordance with this hypothesis that opinions therefore cannot be labelled as real or fake, opinion-based or highly subjective news articles present in the two datasets used in this work could therefore be considered mislabelled samples. These mislabelled samples pose the risk that models may learn incorrect patterns in the data, which would impair, in particular, their ability to generalise \cite{hsu2019niesr}. Lallich et al. find that removing mislabeled classes from training data improved classification on a range of datasets and so this study therefore explores the effect of implementing an additional classification step in which opinion-based articles are identified and removed from the data, creating a filtered training dataset \cite{lallich2002improving}. 

Moreover, similarly to opinion-based news stories, fake news articles often use emotive and sensationalised language \cite{shu2019beyond}. While this language is more commonly found in fake news, it is not exclusive to this class and so by removing opinion-based articles we aim to prevent models unintentionally learning to classify articles based on their degree of sensationalism. 

While a number of studies have been carried out attempting to score sentence subjectivity using datasets of labelled short sentences \cite{riloff2003learning, yu2003towards}, to date no datasets have been found collating fact and opinion-based news stories, which tend to be significantly longer. A new dataset has therefore been created as part of this work, containing 50 rows of data taken from online news sources, of which 25 are factual and 25 are opinion-based. This dataset is publicly available\footnote{\href{https://github.com/CiaraBee/fake_news_classification}{https://github.com/CiaraBee/fake\_news\_classification}}.

Out of the three transformer architectures implemented in this work, DistilBERT achieved the best results on the task of fact vs opinion classification with an accuracy of 78\% on this small dataset. This model is implemented prior to fake news classification, with articles identified as opinion-based then removed from the training dataset. 

\subsection{Methods}
To assess the performance of the models outlined above on fake news classification, four key metrics have been recorded: accuracy, precision, recall and the $F_{1}$ score. Each baseline and transformer model has been trained on 80\% of each dataset, with 10\% used for validation and 10\% reserved for testing. All transformer implementation is done in \emph{PyTorch} \cite{pytorch}, with AdamW \cite{kingma2014adam} providing the best optimization ($\beta_{1} = 0.9$, $\beta_{2} = 0.999$, $\epsilon = 1e-8$). Experimentation has been carried out using a Tesla P100 GPU with an Intel(R) Xeon(R) CPU @ 2.00GHz processor \cite{colabs}.

\section{Experimental Results}
Experiments have been run using the models outlined in section \ref{sec:approach}, testing both in distribution and out of distribution generalisation by evaluating each model's performance on both holdout test sets with similar distributions to training data, and completely unseen datasets with varying topics, styles and class distributions. Following this, the two-step classification pipeline introduced in this study is implemented to compare generalisation performance with and without this additional step and assess its suitability and efficacy to this use case.

\subsection{In Distribution Generalisation}
The baseline and BERT based models have each been trained on the ISOT and CC datasets and evaluated on their holdout test sets to allow comparison between the performance of machine learning, deep learning and transformer based approaches. Table \ref{table:test_set_results_ISOT} shows the results of training and testing on the ISOT dataset, reporting accuracy, precision, recall and $F_1$ scores and Table \ref{table:test_set_results_CC} shows these evaluation metrics when trained and tested on the CC dataset.

\begin{table}[!t]
	\centering
		{\tabulinesep=0mm
			\begin{tabu}{@{\extracolsep{5pt}}l c c c c@{}}
				\hline\hline
				\multicolumn{1}{l}{\multirow{2}{*}{Model}} & 
				\multicolumn{4}{c}{ISOT Dataset}\T\B\\
				\cline{2-5}
									& Accuracy	& Precision & Recall & F\textsubscript{1} Score\T\B\\
\hline\hline

Logistic Regression 				& 0.977 &	0.977 &	0.976 &	0.977\T\\
Na\"ive Bayes 				        & 0.939 &	0.938 &	0.940 &	0.939\\
Random Forest 		            	& 0.957 &	0.959 &	0.954 &	0.956\\
LSTM 			                    & 0.969 &	0.969 &	0.969 &	0.969\\
DistilBERT                          & 0.989 &	0.989 &	0.989 &	0.989\\
BERT                                & 0.990 &	0.990 &	0.989 &	0.989\\
deBERTa 				& \textbf{0.997} &	\textbf{0.997} &	\textbf{0.997} &	\textbf{0.989}\B\\

\hline
\hline
    \end{tabu}
}
\captionsetup[table]{skip=7pt}
\captionof{table}{Comparison results of baseline and BERT based transformer models when trained on 80\% of the ISOT dataset and tested on 10\%.}
\label{table:test_set_results_ISOT}
\end{table}


 


\begin{table}[!t]
	\centering
		{\tabulinesep=0mm
			\begin{tabu}{@{\extracolsep{5pt}}l c c c c@{}}
				\hline\hline
				\multicolumn{1}{l}{\multirow{2}{*}{Model}} & 
				\multicolumn{4}{c}{Combined Corpus Dataset}\T\B\\
				\cline{2-5}
									& Accuracy	& Precision & Recall & F\textsubscript{1} Score\T\B\\
\hline\hline

 Logistic Regression &	0.962 &	0.961 &	0.962 &	0.962\T\\

 Na\"ive Bayes  &	0.866 &	0.871 &	0.864 &	0.865 \\

 Random Forest  &	0.927 &	0.927 &	0.926 &	0.926 \\

  LSTM &	0.942 &	0.944 &	0.941 &	0.942 \\

 DistilBERT  &	0.983 &	0.983 &	0.983 &	0.983 \\

 BERT  &	0.980 &	0.980 &	0.981 &	0.981 \\

 deBERTa &	\textbf{0.988} &	\textbf{0.988} &	\textbf{0.988} &	\textbf{0.988}\B\\

 \hline
\hline
    \end{tabu}
}
\captionsetup[table]{skip=7pt}
\captionof{table}{Comparison results of baseline and BERT based transformer models when trained on 80\% of the CC dataset and tested on 10\%.}
\label{table:test_set_results_CC}

\end{table}

Table \ref{table:test_set_results_ISOT} and \ref{table:test_set_results_CC} show that the three BERT based models outperform the baseline models across all evaluation metrics, with deBERTa scoring the highest across both datasets. The results for BERT and DistilBERT are very similar, with BERT marginally outperforming its more compact counterpart on the ISOT dataset, and DistilBERT performing best on the CC dataset. This indicates that, while smaller, DistilBERT retains much of the natural language understanding of its larger equivalent, performing equally as well in this context but at a much faster speed (60\% faster at inference \cite{sanh2019distilbert}), making it a strong candidate for the task of news classification. These results clearly indicate that transformers are able to learn meaningful patterns and generalise well to datasets with similar distributions. However, with large amounts of online news, including fake news, being published every day by a wide range of sources, it is important to test the robustness of these models when faced with data of varying content, length and writing style.

\subsection{Out of Distribution Generalisation}
In order to assess whether transformers are able to generalise to unseen data with different distributions, experiments have been run in which each model has been trained on either the ISOT or CC dataset and then tested on the other. The results of these experiments are reported in Tables \ref{table:generalisation_results_ISOT} and \ref{table:generalisation_results_CC}. 

\begin{table}[!t]
	\centering
		{\tabulinesep=0mm
			\begin{tabu}{@{\extracolsep{5pt}}l c c c c@{}}
				\hline\hline
				\multicolumn{1}{l}{\multirow{2}{*}{Model}} & 
				\multicolumn{4}{c}{Trained on ISOT Dataset}\T\B\\
				\cline{2-5}
									& Accuracy	& Precision & Recall & F\textsubscript{1} Score\T\B\\
\hline\hline


 Logistic Regression &	0.641 &	0.678 &	0.649 &	0.612\T\\

 Na\"ive Bayes &	0.642 &	0.689 &	0.647 &	0.623\\

 Random Forest &	0.645 &	0.679 &	0.649 &	0.631 \\

  LSTM &	0.632 &	0.746 &	0.639 &	0.691  \\

 DistilBERT &	0.670 &	0.716 &	0.674 &	0.655  \\

 BERT &	0.670 &	\textbf{0.748} &\textbf{0.715} &	\textbf{0.702} \\

 deBERTa &	\textbf{0.697} &	0.702 &	0.699 &	0.695\B\\

\hline
\hline
    \end{tabu}
}
\captionsetup[table]{skip=7pt}
\captionof{table}{Comparison results of baseline and BERT based transformer models
when trained on the ISOT dataset and tested on the Combined Corpus dataset.
}
\label{table:generalisation_results_ISOT}

\end{table}


\begin{table}[!t]
	\centering
		{\tabulinesep=0mm
			\begin{tabu}{@{\extracolsep{5pt}}l c c c c@{}}
				\hline\hline
				\multicolumn{1}{l}{\multirow{2}{*}{Model}} & 
				\multicolumn{4}{c}{Trained on CC Dataset}\T\B\\
				\cline{2-5}
									& Accuracy	& Precision & Recall & F\textsubscript{1} Score\T\B\\
\hline\hline



 Logistic Regression &	0.726 &	0.789 &	0.736 &	0.737\T\\

 Na\"ive Bayes &	0.665 &	0.680 &	0.642 &	0.635 \\

 Random Forest & 0.707 &	0.721 &	0.689 &	0.688 \\

  LSTM &	0.681 &	0.754 &	0.649 &	0.629 \\

 DistilBERT & \textbf{0.775} &	\textbf{0.846} &	\textbf{0.751} &	\textbf{0.750} \\

 BERT &	0.670 &	0.755 &	0.637 &	0.610 \\

 deBERTa &		0.730 &	0.800 &	0.703 &	0.695\B\\
\hline
\hline
    \end{tabu} 

\captionsetup[table]{skip=7pt}
\captionof{table}{Comparison results of baseline and BERT based transformer models
when trained on the Combined Corpus dataset and tested on the ISOT dataset.}
\label{table:generalisation_results_CC}
}

\end{table}

Tables \ref{table:generalisation_results_ISOT} and \ref{table:generalisation_results_CC} show a clear drop in performance when compared with the results of tables \ref{table:test_set_results_ISOT} and \ref{table:test_set_results_CC}. However, as these datasets are from different sources, this is to be expected since a difference in the data distribution between the training and test sets always leads to a drop in performance. Nonetheless, these results show that BERT based models consistently achieve better performance than the baseline comparisons across all evaluation metrics. A major advantage of transformers for this task is their pre-training on large and varied datasets which Hendrycks et al. suggest improves robustness and generalisation \cite{hendrycks2019using}.

Table \ref{table:generalisation_results_CC} shows a generally stronger generalisation performance across models, potentially due to the larger size and broader topic scope of the CC dataset. Moreover, these results show slightly more variation across datasets and models, with the logistic regression classifier achieving higher recall and $F_{1}$ scores than BERT and deBERTa. 

While the transformers perform well overall on this task, it is clear that there are more incorrect predictions being made compared to in distribution generalisation. As the values reported in Tables \ref{table:class_results_ISOT} and \ref{table:class_results_CC} are macro averages, we can better understand these results by investigating the evaluation metrics for each class.


\begin{table*}[!t]
	\centering
		{\tabulinesep=0mm
			\begin{tabu}{@{\extracolsep{5pt}}l c c c c c c c c c@{}}
				\hline\hline
				\multicolumn{1}{l}{\multirow{2}{*}{Model}} & 
				\multicolumn{2}{c}{Accuracy}
				&
				\multicolumn{2}{c}{Precision}
				& 
				\multicolumn{2}{c}{Recall}
				& F\textsubscript{1} Score\T\B \\
				\cline{2-9}
				& Fake & Real & Fake & Real & Fake & Real & Fake & Real\T\B \\

\hline\hline



 \hline
 DistilBERT &0.670 & 0.670 & 0.611 & 0.821 & 0.896 & \textbf{0.453} & 0.726 & 0.583\T \\

 BERT &	0.711 & 0.711 & 0.646 & 0.849 & 0.903 & 0.527 & 0.753 & 0.651 \\

 deBERTa &	0.697 & 0.697 & 0.662 & 0.742 & 0.774 & 0.622 & 0.714 & 0.677\B \\
 \hline
 \hline

\end{tabu}
}
 \captionsetup[table]{skip=2pt}
\captionof{table}{Results of generalisation experiments for transformers trained on the ISOT dataset and tested on the CC dataset for each class.}
 \label{table:class_results_ISOT}

\end{table*}



\begin{table*}[!t]
	\centering
		{\tabulinesep=0mm
			\begin{tabu}{@{\extracolsep{5pt}}l c c c c c c c c c@{}}
				\hline\hline
				\multicolumn{1}{l}{\multirow{2}{*}{Model}} & 
				\multicolumn{2}{c}{Accuracy}
				&
				\multicolumn{2}{c}{Precision}
				& 
				\multicolumn{2}{c}{Recall}
				& F\textsubscript{1} Score\T\B \\
				\cline{2-9}
				& Fake & Real & Fake & Real & Fake & Real & Fake & Real\T\B \\

\hline\hline


 DistilBERT &0.775 & 0.775	 &  0.980 & 0.712 & 0.510 & 0.991 & 0.671 & 0.829\T \\

 BERT &	0.670 & 0.670 & 0.879 & 0.631 & \textbf{0.308} & 0.965 & \textbf{0.456} & 0.763 \\

 deBERTa &	0.730 & 0.730 & 0.923 & 0.678 & \textbf{0.435} & 0.970 & 0.591 & 0.798\B \\
 \hline
 \hline

\end{tabu}
}
\captionsetup[table]{skip=7pt}
\captionof{table}{ Results of generalisation experiments for transformers trained on the
CC dataset and tested on the ISOT dataset for each class.}
\label{table:class_results_CC}
        \end{table*}


These results show that while the BERT based transformers generalise well overall, there are large differences in the precision and recall values for each class, indicating skews towards predicting a certain class. Of particular note are the evaluation metrics below 0.5, which indicate performance lower than random chance and suggest that these models may be learning spurious connections from the data that significantly hinder generalisation to unseen data. In order to investigate whether mislabelled subjective samples may be the cause of such incorrect patterns, we go on to assess the effect of implementing the two-stage classification pipeline demonstrated in Figure \ref{fig:pipeline_diagram}.

\subsubsection{Filtering Opinion-Based Articles}
As outlined in Figure \ref{fig:pipeline_diagram}, a filtered training dataset has been created. All baseline and BERT based transformers have been trained on this filtered training data and evaluated against the alternative dataset to investigate how this additional step affects out of distribution generalisation. The results from the two-step classification pipeline are compared with the one-step classification in Tables \ref{table:step_1_results_ISOT} and \ref{table:step_1_results_CC} below.
    

\begin{table*}[!t]
	\centering
		{\tabulinesep=0mm
			\begin{tabu}{@{\extracolsep{5pt}}l c c c c c c c c c@{}}
				\hline\hline
				\multicolumn{1}{l}{\multirow{2}{*}{Model}} & 
				\multicolumn{2}{c}{Accuracy}
				&
				\multicolumn{2}{c}{Precision}
				& 
				\multicolumn{2}{c}{Recall}
				& F\textsubscript{1} Score\T\B \\
				\cline{2-9}
				& Two Step & One Step & Two Step & One Step & Two Step & One Step & Two Step & One Step\T\B \\

\hline\hline
 Logistic Regression &	0.662 &	0.641 &	0.747 &	0.678 &	0.669 &	0.649 &	0.636 &	0.612\T \\

 Na\"ive Bayes &	0.630 &	0.642 &	0.634 &	0.689 &	0.632 &	0.647 &	0.629 &	0.623 \\

 Random Forest &	0.639 &	0.645 &	0.653 &	0.679 &	0.642 &	0.649 &	0.634 &	0.631 \\

  LSTM &	0.632 &	0.632 &	0.670 &	0.746 &	0.637 &	0.639 &	0.616 &	0.691 \\

 DistilBERT &	0.678 &	0.670 &	0.722 &	0.716 &	0.682 &	0.674 &	0.713 &	0.655\\

 BERT &	\textbf{0.710} & 0.670 &	0.740 &	\textbf{0.748} &	0.715 &	\textbf{0.715} &	0.733 &	\textbf{0.702} \\

 deBERTa &	0.674 &	0.697 &	0.675 &	0.702 &	0.673 &	0.699 &	0.675 &	0.695\B \\
\hline
\hline

\end{tabu}
}
 \captionsetup[table]{skip=7pt}
\captionof{table}{Comparison results showing generalisation performance of all models with the two step classification pipeline applied and without when trained on the ISOT dataset and tested on the CC dataset.}
        \label{table:step_1_results_ISOT}
        \end{table*}

Tables \ref{table:step_1_results_ISOT} and \ref{table:step_1_results_CC} clearly show a difference in the effectiveness of removing opinion-based articles between datasets, with models trained on the CC dataset seeing improvements from their removal, while those trained on the ISOT dataset overall do not. It is important to note that in removing opinion-based articles from the data, the size of the resulting dataset is reduced. The LSTMs and transformers, being neural networks, have a large number of parameters and so \enquote{require large amounts of data for training in order for over-fit avoidance and better model generalisation} \cite{guo2019augmenting}. The results shown in Table \ref{table:step_1_results_ISOT} suggest that the resulting dataset after removing opinion-based articles is too small for these models to generalise well, resulting in generally reduced performances.


\begin{table*}[!t]
	\centering
		{\tabulinesep=0mm
			\begin{tabu}{@{\extracolsep{5pt}}l c c c c c c c c c@{}}
				\hline\hline
				\multicolumn{1}{l}{\multirow{2}{*}{Model}} & 
				\multicolumn{2}{c}{Accuracy}
				&
				\multicolumn{2}{c}{Precision}
				& 
				\multicolumn{2}{c}{Recall}
				& F\textsubscript{1} Score\T\B \\
				\cline{2-9}
				& Two Step & One Step & Two Step & One Step & Two Step & One Step & Two Step & One Step\T\B \\

\hline\hline


 Logistic Regression &	0.769 &	0.726 &	0.808 &	0.789 &	0.749 &	0.736 &	0.750 &	0.737\T \\

 Na\"ive Bayes &	0.631 &	0.665 &	0.757 &	0.680 &	0.591 &	0.642 &	0.535 &	0.635 \\

 Random Forest &	0.694 &	0.707 &	0.721 &	0.721 &	0.672 &	0.689 &	0.667 &	0.688 \\
 
  LSTM &	0.736 &	0.681 &	0.760 &	0.745 &	0.718 &	0.649 &	0.718 &	0.629 \\

 DistilBERT &	0.714 &	0.681 &	0.735 &	0.739 &	0.729 &	0.652 &	0.713 &	0.693 \\

 BERT &	0.667 &	0.670 &	0.629 &	0.755 &	0.699 &	0.638 &	0.638 &	0.672 \\

 deBERTa &	\textbf{0.808} &	0.730 &	\textbf{0.839} &	0.800 &	\textbf{0.792} &	0.703 &	\textbf{0.796} &	0.695\B \\
 \hline
 \hline

\end{tabu}
}
        \captionsetup[table]{skip=2pt}
\captionof{table}{Comparison results showing generalisation performance of all models with the two step classification pipeline applied and without when trained on the CC dataset and tested on the ISOT dataset.}
        \label{table:step_1_results_CC}
        \end{table*}
 

However, when training on the larger more diverse CC dataset, Table \ref{table:step_1_results_CC} shows that the removal of opinion-based articles allows deBERTa to achieve the highest performance across all metrics. DeBERTa achieves particularly large gains in generalisation with this filtering step applied, increasing its $F_{1}$ score on this task by 10.1\%. With the exception of BERT, this opinion filtering step notably increases the generalisation performance of both the transformers and the LSTM.

\begin{figure}[t!]
	\centering
	\includegraphics[width=0.75\linewidth]{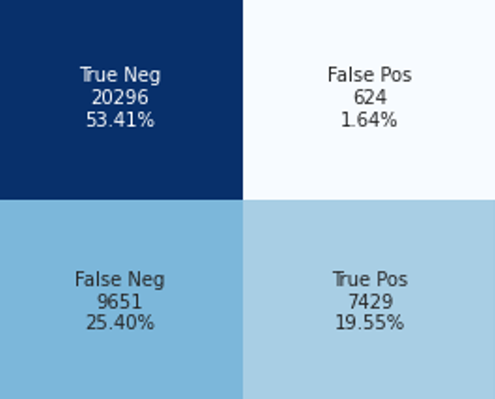}
	\captionsetup[figure]{skip=7pt}
	\captionof{figure}{Confusion matrix for deBERTa model trained on CC dataset and tested on ISOT.}
	\label{fig:deberta_confusion_mat}
\end{figure}

\begin{figure}[t!]
	\centering
	\includegraphics[width=0.75\linewidth]{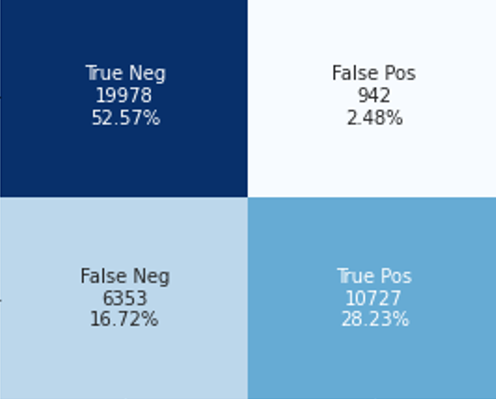}
	\captionsetup[figure]{skip=7pt}
	\captionof{figure}{Confusion matrix for deBERTa model trained on CC dataset and tested on ISOT with two-step pipeline implemented.}
	\label{fig:deberta_step_1_confusion_mat}
\end{figure}

Figure \ref{fig:deberta_confusion_mat} shows the confusion matrix for the deBERTa model trained on the CC dataset and evaluated on the ISOT dataset. These results show that this model is largely skewed towards predicting negative values, with a high proportion of false negative and a low proportion of true positives. Figure \ref{fig:deberta_step_1_confusion_mat} shows the confusion matrix for this same model after implementing the two-step classification pipeline. These results show a smaller proportion of false negative values along with an increased proportion of true positives, indicating that removing opinion-based articles has resulted in better classification and generalisation performance.
\newline

\section{Future Work}

While the dataset used to train the fact/opinion classifier was very small (50 news articles), this additional step nonetheless resulted in notable increases in generalisation performance for models trained on the CC dataset, indicating that opinion filtering may aid the learning process by removing mislabelled samples. There is much scope for developing a larger dataset containing labelled fact and opinion-based news articles which would allow for better performance in identifying opinion-based articles, and thus likely improve generalisation further. 

This study has shown that transformers may learn incorrect patterns from data that harm their out of distribution generalisation performance. Following this, there is further work to be done in quantifying, understanding and eventually preventing transformers learning spurious patterns in data. While this work explores removal of mislabelled data to tackle this problem, data augmentation has also been suggested as an approach to improve the robustness of models
\cite{shorten2021text}. 

\section{Conclusion}
The digitisation of media, and in particular social media, has allowed news, both real and fake, to propagate faster than ever before \cite{figueira2017current}. Auditing the veracity of news content posted online at the earliest point of detection possible is therefore crucial in tackling the fake news problem. Online news as its core is simply text, and so this study assesses the effectiveness of using using transformers, the state-of-the-art in natural language processing, to classify news based solely on textual content, paying particular attention to out of distribution generalisation.

This study has shown that transformers such as BERT, DistilBERT and deBERTa outperform machine learning and deep learning baseline alternatives (logistic regression, na\"ive Bayes, random forest and LSTM classifiers) in news classification when testing in distribution generalisation and out of distribution generalisation, achieving a peak accuracy of 77.5\% by the DistilBERT model on the latter. Additionally, we have addressed the subjective and inconsistent nature of fake news by proposing a two-step classification pipeline which identifies and removes opinion-based news articles from the training data used by the final news classifier. In doing so, the most subjective and therefore unpredictable samples are filtered out of the data to prevent models learning incorrect patterns that do not generalise. This two-step classification process improves the accuracy of deBERTa predictions by 7.8\% to a peak of 80.8\% and improves its $F_1$ score by 10.1\%. However, the effectiveness of this method seems to be largely dependent on the dataset, with larger and more varied datasets producing superior results.

\bibliographystyle{plain}
\bibliography{references}
\end{document}